\documentclass{article}

\usepackage{microtype}
\usepackage{graphicx}
\usepackage{subfigure}
\usepackage{booktabs} 

\usepackage{hyperref}

\usepackage[accepted]{icml2025}

\usepackage{amsmath}
\usepackage{amssymb}
\usepackage{mathtools}
\usepackage{amsthm}

\usepackage[capitalize,noabbrev]{cleveref}

\theoremstyle{plain}
\newtheorem{theorem}{Theorem}

\theoremstyle{definition}
\newtheorem{definition}[theorem]{Definition}

\theoremstyle{remark}

\usepackage[textsize=tiny]{todonotes}

\icmltitlerunning{GTED: A Faithful Evaluation Metric for Statement Autoformalization}

\definecolor{linkcolor}{RGB}{255,0,0}
\definecolor{urlcolor}{RGB}{255,105,180}
\definecolor{citecolor}{RGB}{66,168,235}
\usepackage{graphicx}
\usepackage{booktabs}
\usepackage{xspace}
\usepackage{colortbl}

\usepackage{multirow}
\usepackage{booktabs}
\usepackage{makecell}   
\usepackage{array}
\usepackage{listings}
\newcommand{\smallsec}[1]{\paragraph{#1.}}
\colorlet{punct}{red!60!black}
\definecolor{background}{HTML}{EEEEEE}
\definecolor{delim}{RGB}{20,105,176}
\colorlet{keyword}{magenta!60!black}
\lstdefinelanguage{lean}{
    basicstyle=\footnotesize\ttfamily,
    numbers=left,
    numberstyle=\scriptsize,
    stepnumber=1,
    numbersep=8pt,
    showstringspaces=false,
    breaklines=true,
    frame=lines,
    backgroundcolor=\color{background},
    literate=
     *{axiom}{{{\color{keyword}axiom}}}{5}
      {theorem}{{{\color{keyword}theorem}}}{7}
      {lemma}{{{\color{keyword}lemma}}}{5}
      {by}{{{\color{keyword}by}}}{2}
      {Lemma}{{{\color{keyword}Lemma}}}{5}
      {Proof}{{{\color{keyword}Proof}}}{5}
      {Qed}{{{\color{keyword}Qed}}}{3}
      {:}{{{\color{punct}{:}}}}{1}
      {,}{{{\color{punct}{,}}}}{1}
      {\{}{{{\color{delim}{\{}}}}{1}
      {\}}{{{\color{delim}{\}}}}}{1}
      {[}{{{\color{delim}{[}}}}{1}
      {]}{{{\color{delim}{]}}}}{1}
      {ℕ}{{\ensuremath{\mathbb{N}}}}{1}
      {ℤ}{{\ensuremath{\mathbb{Z}}}}1
      {ℝ}{{\ensuremath{\mathbb{R}}}}{1}
      {ℚ}{{\ensuremath{\mathbb{Q}}}}1
      {ℂ}{{\ensuremath{\mathbb{C}}}}1
      {∩}{{\ensuremath{\cap}}}1
    {∪}{{\ensuremath{\cup}}}1
    {⊂}{{\ensuremath{\subseteq}}}1
    {⊆}{{\ensuremath{\subseteq}}}1
    {⊄}{{\ensuremath{\nsubseteq}}}1
    {⊈}{{\ensuremath{\nsubseteq}}}1
    {⊃}{{\ensuremath{\supseteq}}}1
    {⊇}{{\ensuremath{\supseteq}}}1
    {⊅}{{\ensuremath{\nsupseteq}}}1
    {⊉}{{\ensuremath{\nsupseteq}}}1
    {∈}{{\ensuremath{\in}}}1
    {∉}{{\ensuremath{\notin}}}1
    {∋}{{\ensuremath{\ni}}}1
    {∌}{{\ensuremath{\notni}}}1
    {∅}{{\ensuremath{\emptyset}}}1
    {∫}{{\ensuremath{\int}}}1
    {∑}{{\ensuremath{\mathrm{\Sigma}}}}1
    {Π}{{\ensuremath{\mathrm{\Pi}}}}1
    {≤}{{\ensuremath{\leq}}}1
    {≥}{{\ensuremath{\geq}}}1
    {≠}{{\ensuremath{\neq}}}1
    {≈}{{\ensuremath{\approx}}}1
    {≡}{{\ensuremath{\equiv}}}1
    {≃}{{\ensuremath{\simeq}}}1
    {α}{{\ensuremath{\mathrm{\alpha}}}}1
    {β}{{\ensuremath{\mathrm{\beta}}}}1
    {γ}{{\ensuremath{\mathrm{\gamma}}}}1
    {δ}{{\ensuremath{\mathrm{\delta}}}}1
    {ε}{{\ensuremath{\mathrm{\varepsilon}}}}1
    {ζ}{{\ensuremath{\mathrm{\zeta}}}}1
    {η}{{\ensuremath{\mathrm{\eta}}}}1
    {θ}{{\ensuremath{\mathrm{\theta}}}}1
    {ι}{{\ensuremath{\mathrm{\iota}}}}1
    {κ}{{\ensuremath{\mathrm{\kappa}}}}1
    {μ}{{\ensuremath{\mathrm{\mu}}}}1
    {ν}{{\ensuremath{\mathrm{\nu}}}}1
    {ξ}{{\ensuremath{\mathrm{\xi}}}}1
    {π}{{\ensuremath{\mathrm{\mathnormal{\pi}}}}}1
    {ρ}{{\ensuremath{\mathrm{\rho}}}}1
    {σ}{{\ensuremath{\mathrm{\sigma}}}}1
    {τ}{{\ensuremath{\mathrm{\tau}}}}1
    {φ}{{\ensuremath{\mathrm{\varphi}}}}1
    {χ}{{\ensuremath{\mathrm{\chi}}}}1
    {ψ}{{\ensuremath{\mathrm{\psi}}}}1
    {ω}{{\ensuremath{\mathrm{\omega}}}}1
    {Γ}{{\ensuremath{\mathrm{\Gamma}}}}1
    {Δ}{{\ensuremath{\mathrm{\Delta}}}}1
    {Θ}{{\ensuremath{\mathrm{\Theta}}}}1
    {Λ}{{\ensuremath{\mathrm{\Lambda}}}}1
    {Σ}{{\ensuremath{\mathrm{\Sigma}}}}1
    {Φ}{{\ensuremath{\mathrm{\Phi}}}}1
    {Ξ}{{\ensuremath{\mathrm{\Xi}}}}1
    {Ψ}{{\ensuremath{\mathrm{\Psi}}}}1
    {Ω}{{\ensuremath{\mathrm{\Omega}}}}1
    {↦}{{\ensuremath{\mapsto}}}1
    {←}{{\ensuremath{\leftarrow}}}1
    {<-}{{\ensuremath{\leftarrow}}}1
    {→}{{\ensuremath{\rightarrow}}}1
    {↔}{{\ensuremath{\leftrightarrow}}}1
    {⇒}{{\ensuremath{\Rightarrow}}}1
    {⟹}{{\ensuremath{\Longrightarrow}}}1
    {⇐}{{\ensuremath{\Leftarrow}}}1
    {⟸}{{\ensuremath{\Longleftarrow}}}1
    {Σ}{{\ensuremath{\Sigma}}}1
    {Π}{{\ensuremath{\Pi}}}1
    {∀}{{\ensuremath{\forall}}}1
    {∃}{{\ensuremath{\exists}}}1
    {λ}{{\ensuremath{\mathrm{\lambda}}}}1
    {∧}{{\ensuremath{\wedge}}}1
    {∨}{{\ensuremath{\vee}}}1
    {¬}{{\ensuremath{\neg}}}1
    {⊢}{{\ensuremath{\vdash}}}1
    {‖}{{\ensuremath{\|}}}1
    {₁}{{\ensuremath{_1}}}1
    {₂}{{\ensuremath{_2}}}1
    {₃}{{\ensuremath{_3}}}1
    {₄}{{\ensuremath{_4}}}1
    {₅}{{\ensuremath{_5}}}1
    {₆}{{\ensuremath{_6}}}1
    {₇}{{\ensuremath{_7}}}1
    {₈}{{\ensuremath{_8}}}1
    {₉}{{\ensuremath{_9}}}1
    {₀}{{\ensuremath{_0}}}1
    {ᵢ}{{\ensuremath{_i}}}1
    {ⱼ}{{\ensuremath{_j}}}1
    {ₐ}{{\ensuremath{_a}}}1
    {⁻¹}{{\ensuremath{^{-1}}}}1
    {¹}{{\ensuremath{^1}}}1
    {ₙ}{{\ensuremath{_n}}}1
    {ₘ}{{\ensuremath{_m}}}1
    {ₚ}{{\ensuremath{_p}}}1
    {↑}{{\ensuremath{\uparrow}}}1
    {↓}{{\ensuremath{\downarrow}}}1
    {⊥}{{\ensuremath{\perp}}}1
    {∞}{{\ensuremath{\infty}}}1
    {∂}{{\ensuremath{\partial}}}1
    {√}{{\ensuremath{\sqrt}}}1
    {∘}{{\ensuremath{\circ}}}1
    {×}{{\ensuremath{\times}}}1
    {∆}{{\ensuremath{\triangle}}}1
    {⟨}{{\ensuremath{\langle}}}1
    {⟩}{{\ensuremath{\rangle}}}1
    {ℒ}{{\ensuremath{\mathcal{L}}}}1,
}
\lstdefinestyle{appendixstyle}{
    language=lean,
    numbers=none, 
    frame=none,
    backgroundcolor=\color{white}, 
}
\usepackage{enumitem}
\usepackage{tcolorbox}
\usepackage{pifont}
\usepackage{framed}
\usepackage{amsmath}

\definecolor{authorAcolor}{RGB}{206,0,16}  
\definecolor{authorBcolor}{RGB}{133,153,0}  
\definecolor{authorCcolor}{RGB}{38,139,210} 
\definecolor{authorDcolor}{RGB}{255,140,0}  
\definecolor{authorEcolor}{RGB}{108,113,196} 
\definecolor{authorFcolor}{RGB}{42,161,152}

\usepackage[all]{xy}
\usepackage{mathtools}
\usepackage{extarrows}

\begin{document}

\twocolumn[
\icmltitle{Generalized Tree Edit Distance (GTED): A Faithful Evaluation Metric for Statement Autoformalization}



\icmlsetsymbol{equal}{*}

\begin{icmlauthorlist}
\icmlauthor{Yuntian Liu}{1,equal}
\icmlauthor{Tao Zhu}{1,equal}
\icmlauthor{Xiaoyang Liu}{1,equal}
\icmlauthor{Yu Chen}{1}
\icmlauthor{Zhaoxuan Liu}{1}
\icmlauthor{Qingfeng Guo}{1}
\icmlauthor{Jiashuo Zhang}{1}
\icmlauthor{Kangjie Bao}{1}
\icmlauthor{Tao Luo}{1,2,3}
\end{icmlauthorlist}

\icmlaffiliation{1}{School of Mathematical Sciences, Shanghai Jiao Tong University}
\icmlaffiliation{2}{Institute of Natural Sciences, MOE-LSC, Shanghai Jiao Tong University}
\icmlaffiliation{3}{CMA-Shanghai, Shanghai Artificial Intelligence Laboratory}

\icmlcorrespondingauthor{Tao Luo}{luotao41@sjtu.edu.cn}

\icmlkeywords{Evaluation Metric, Autoformalization, Lean 4, Operator Tree}
\vskip 0.3in
]

\printAffiliationsAndNotice{\icmlEqualContribution} 

\begin{abstract}
Statement autoformalization, the automated translation of statements from natural language into formal languages, has become a subject of extensive research, yet the development of robust automated evaluation metrics remains limited.
Existing evaluation methods often lack semantic understanding, face challenges with high computational costs, and are constrained by the current progress of automated theorem proving.
To address these issues, we propose \textbf{GTED} (\textit{\underline{G}eneralized \underline{T}ree \underline{E}dit \underline{D}istance}), a novel evaluation framework that first standardizes formal statements and converts them into operator trees, then determines the semantic similarity using the eponymous GTED metric.
Across the miniF2F and ProofNet benchmarks, GTED consistently ranks as a top-performing metric, achieving the highest accuracy and Kappa on miniF2F and the joint-highest accuracy on ProofNet.
This strong overall performance provides the community with a computationally lightweight and more faithful metric for automated evaluation.
The code and experimental results are available at \url{https://github.com/XiaoyangLiu-sjtu/GTED}.
\end{abstract}

\section{Introduction} \label{sec:introduction}
Formal languages like Isabelle \citep{isabelle}, HOL Light \citep{hol_light}, Coq \citep{coq}, and Lean \citep{lean4_2015, lean4_2021} have recently garnered significant attention within the mathematical community due to their utility in rigorously verifying the correctness of proofs. Nevertheless, the endeavor of formalizing mathematical content demands substantial time and effort, alongside a profound familiarity with these specialized languages, rendering the process inherently labor-intensive. Therefore, the autoformalization task \citep{autoformalization_definition}, defined as translating theorem statements and proofs from natural language into their formal language counterparts, has become a subject of extensive research.

However, despite rapid advancements in autoformalization, the development of robust automated evaluation metrics remains notably limited. Existing evaluation metrics each exhibits specific limitations. For instance, syntax-based methods, such as Typecheck \citep{mma}, validate compliance with formal grammar but often overlook semantic content. Text similarity metrics, like BLEU \citep{bleu}, are overly sensitive to lexical variation, failing to capture deeper structural or logical equivalence. While proof-based approaches \citep{beq} offer logical rigor, they frequently miss valid formalizations due to the limited capabilities of the underlying prover. Concurrently, LLM-assisted evaluations \citep{lean_workbook} contend with reproducibility and high cost concerns. These limitations collectively highlight a significant gap: the absence of a consistent, flexible, and efficient semantic evaluation metric.

In this work, we propose GTED (\textit{Generalized Tree Edit Distance}), a novel evaluation framework based on tree edit distance \citep{ted}. Our framework follows a three-stage process. First, it standardizes formal statements by leveraging the Lean Language Server. Next, it constructs operator trees (OPTs) by parsing the syntactic structure of each statement. Finally, it enhances the tree edit distance calculation to compute a similarity score. Rather than outputting a binary decision, our framework yields a continuous similarity score in the range [0,1], enabling a more nuanced and interpretable evaluation of semantic similarity.

We demonstrate the superiority of the proposed GTED metric by evaluating it against a suite of baseline metrics on miniF2F and ProofNet. On miniF2F, GTED surpasses all baselines, securing the top accuracy (70.73\%) and Kappa score (0.438). On ProofNet, it achieves the joint-highest accuracy (69.89\%) and a highly competitive Kappa score (0.402). Crucially, unlike metrics that are either too strict (e.g., Identity Match, with low recall) or too lenient (e.g., Typecheck, with low precision), GTED also provides the best balance between precision and recall, establishing a more reliable and robust measure for semantic alignment.

\begin{figure*}[ht]
\centering
\includegraphics[width=2\columnwidth]{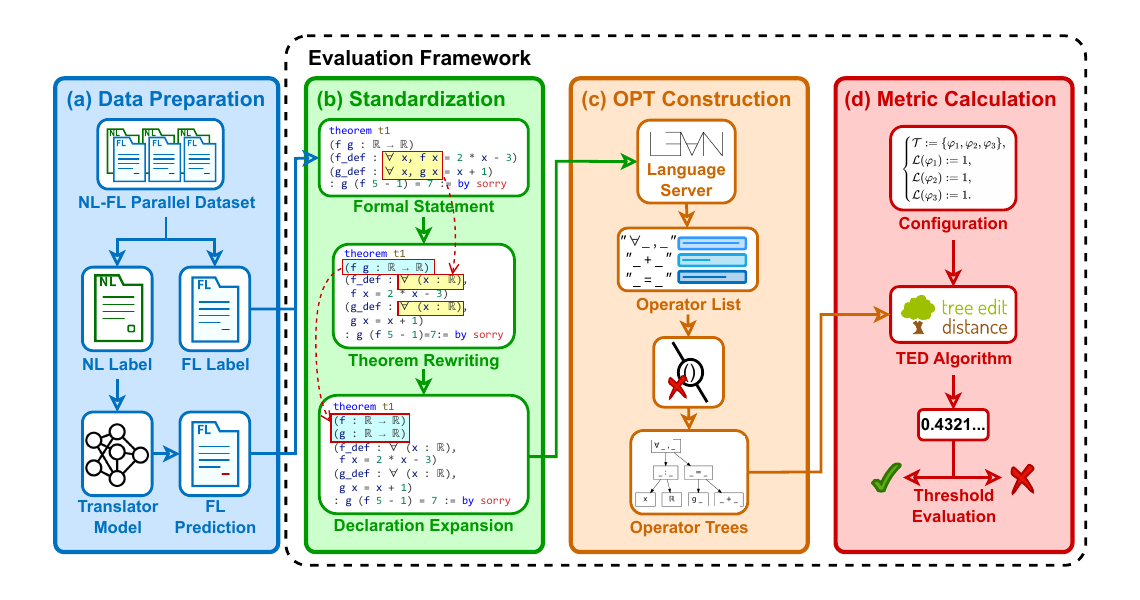}
\caption{Illustration of GTED (\textit{Generalized Tree Edit Distance}) for statement formalization. (a) Data Preparation: A translator model generates a formal language (FL) prediction from a natural language (NL) input. (b) Standardization: Both the FL prediction and the ground-truth FL label are standardized via theorem rewriting and variable expansion. (c) OPT Construction: The standardized statements are then parsed into operator trees (OPTs) using the Lean Language Server. (d) Metric Calculation: Finally, the GTED is computed between the two OPTs to yield the similarity score.}
\label{fig:framework}
\end{figure*}

Our main contributions are as follows:
\begin{itemize}[topsep=0pt, itemsep=0pt, parsep=0pt]
\item[1.] We propose GTED (\textit{Generalized Tree Edit Distance}), a novel evaluation metric that introduces and generalizes the standard tree edit distance for the semantic evaluation of statement autoformalization.
\item[2.] We implement a procedure to convert formal statements into operator tree representations by leveraging the Lean Language Server.
\item[3.] We demonstrate through extensive experiments that GTED achieves both the highest accuracy and the strongest alignment with human expert judgment on miniF2F, and obtains the joint-highest accuracy and on ProofNet.
\end{itemize}

\section{Related Work} \label{sec:related_work}
\smallsec{Autoformalization}
Research in autoformalization, particularly for theorem statements, currently shows diverse approaches. While early methods \citep{nmt_1, nmt_2} often employ neural machine translation, the transformative progress of Large Language Models (LLMs) now leads to the emergence of three dominant strategies for LLM-based autoformalization. These include exploring the efficacy of few-shot prompting \citep{llm_icl_1, llm_icl_2, llm_icl_3}; improving capabilities by fine-tuning LLMs \citep{herald, pda, atlas} on relevant parallel statements; and integrating retrieval-augmented generation \citep{rag} to achieve enhanced performance.

\smallsec{Automated Evaluation}
In the realm of automated evaluation, early efforts predominantly employ metrics based on grammatical validity \citep{mma} and string similarity \citep{proofnet}, yet these struggle with semantic understanding. FormalAlign \citep{formal_align} innovatively integrates autoformalization with evaluation by simultaneously generating a formal statement and its corresponding evaluation score. While this represents a significant contribution to the fields of autoformalization and automated evaluation, its scoring mechanism cannot be used as a standalone evaluation metric. Simultaneously, cross-provability \citep{euclidean_geo, symbolic_eq, beq} between formal statements emerges as a widely accepted standard for automated evaluation, but its effectiveness is constrained by the current progress in automated theorem proving.

\smallsec{Operator Trees}
Operator trees (OPTs) are a foundational data structure in mathematical information retrieval (MIR), a field dedicated to the effective retrieval of mathematical formulae from digital corpora \citep{zhong2022evaluating}. The prevalence of OPTs stems from their unique capacity to encode both the two-dimensional symbol layout and the underlying syntactic hierarchy of mathematical notation, which enables robust recognition and semantic understanding \citep{zanibbi2002recognizing, zanibbi2012recognition}. By preserving vital information such as operator precedence and operand relationships, OPTs enable a more nuanced understanding of mathematical expressions. This semantic depth is foundational to leading MIR systems like WikiMirs \citep{gao2016math} and Approach0 \citep{zhong2022applying}, underpinning their ability to perform accurate structural matching and retrieval.

\section{Methodology} \label{sec:methodology}
The methodology follows a three-stage pipeline to compute the similarity between formal statements: Section \ref{sec:syntax_standardization} details the syntax standardization process, Section \ref{sec:opt_construction} describes the construction of operator trees (OPTs) from the normalized statements, and Section \ref{sec:metric_calculation} concludes with the formal mathematical definition and formulation of the GTED metric.

\subsection{Syntax Standardization} \label{sec:syntax_standardization}
To overcome the analytical challenges posed by syntactic variations in formal statements, such as optional type declarations and compact multivariable declarations, and enable precise structural decomposition, we first standardize the formal statements prior to constructing the operator trees. This standardization process employs two key transformations: (1) theorem rewriting and (2) variable expansion.

\smallsec{Theorem Rewriting}
For any given formal statement, the Lean Language Server inherently performs comprehensive syntax normalization as part of its compilation process. This crucial feature handles tasks like type annotation completion, notation expansion, and general syntax standardization. By programmatically interacting with the language server, we effectively rewrite the original formal statement into a processed, normalized form. This resulting representation retains all necessary semantic information while eliminating superficial syntactic variations, thereby preparing it for subsequent operator tree construction. For example, Figure \ref{fig:framework}(b) illustrates the rewriting of the original formal statement, specifically by adding the crucial type information $\mathbb{R}$ for x.

\smallsec{Variable Expansion}
To standardize variable declaration conventions, we expand compact variable declarations (which allow multiple variables to share a single type annotation) into distinct, individual declarations. This process ensures structural clarity in the resulting operator trees while strictly preserving the original mathematical semantics. For instance, Figure \ref{fig:framework}, part (b) illustrates this by transforming a compact declaration like (f g : $\mathbb{R} \to \mathbb{R}$) into individual declarations (f : $\mathbb{R} \to \mathbb{R}$) and (g : $\mathbb{R} \to \mathbb{R}$).

\subsection{OPT Construction} \label{sec:opt_construction}
\begin{figure}[ht]
\centering
\includegraphics[width=\columnwidth]{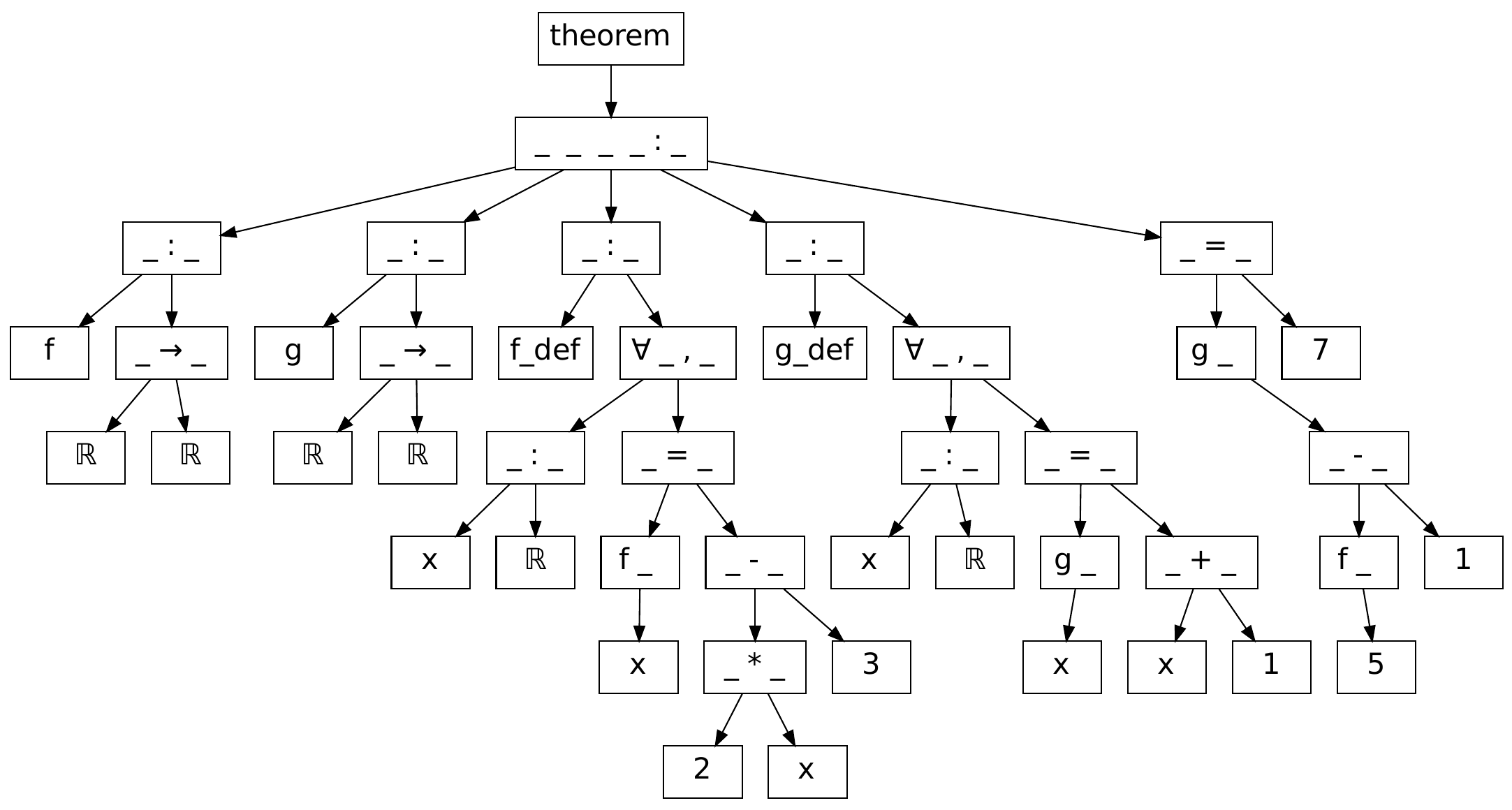}
\caption{Example of an operator tree for a formal statement.}
\label{fig:tree}
\end{figure}

Following standardization, we programmatically interact with the Lean Language Server again to obtain the scope of each element within the formal statement. This scope information is then used to construct an operator tree, integrating two additional operations: (1) placeholder representation and (2) parentheses removal. Figure \ref{fig:tree} visually demonstrates a constructed operator tree.

\smallsec{Placeholder Representation}
For the precise capture of structure during operator tree construction and the efficient reconstruction of the original formal statement, we utilize the placeholder representation. This method outlines the tree's components as follows:
\begin{itemize}[topsep=0pt, itemsep=0pt, parsep=0pt]
\item Non-leaf nodes are assigned their operator and associated parameter slots.
\item Parameter slots are indicated by underscores (\_).
\item Leaf nodes represent mathematical objects.
\end{itemize}
From a long-term perspective, the strategic design of placeholders is crucial because it facilitates the rapid restoration of partial subtrees to their original formal sub-statements. This capability, in turn, enables mutual proof to establish the equivalence of these subtrees. Consequently, this approach seamlessly integrates proof-based evaluation directly into the tree structure. This presents a critical advantage: proving these sub-statements is significantly less challenging than proving the entire statements, thus carving out a key direction for our future work.

\smallsec{Parentheses Removal}
Another crucial aspect of OPT construction involves the deliberate elimination of round-bracket operators. This approach is driven by a key insight: these brackets are primarily an artifact of linear, string-based representations. In such linear forms, explicit punctuation is necessary to rigorously define grouping and operator precedence when constructing new entities from existing ones. Conversely, languages inherently represented as trees naturally and explicitly encode these structural dependencies through their topology. Thus, all superfluous round bracket operators are removed from the tree representation, resulting in a more abstract and structurally precise representation that reduces unnecessary computational overhead for subsequent tree edit distance calculations.

\subsection{Metric Calculation} \label{sec:metric_calculation}


\smallsec{Definitions and Notation}


We begin by introducing the formal definitions and mathematical notation that provide the foundation for our proposed method.
\begin{definition}[Basic Notation] \label{def:operator_tree}
An \textbf{Operator Tree} is a rooted tree, where: 
\begin{itemize}[topsep=0pt, itemsep=0pt, parsep=0pt]
\item Every leaf node is labeled to represent an independent variable or constant in Lean. 
\item Every internal node is labeled to represent an operator in Lean, including (dependent) functions and other objects that take multiple arguments. 
\end{itemize}
\textit{*Throughout this work, we consider only finite trees. }

Denote that $t$ is a \textbf{subtree} of $T$ as $t\subseteq T$. 

Define the \textbf{Quotient Tree} of an operator tree $T$ and its subtree $t$, denoted as $T/t$, as the operator tree that treats $t$ in $T$ as a leaf node. 
\end{definition}

\begin{definition}[Tree Transformations] \label{def:transformation}
A \textbf{Special Tree Transformation} is defined as a pair of operator trees $f=(T_1,T_2)$, and we denote $f(T_1)=T_2$. Denote the collection of special tree transformations as $\mathcal{S}$ ($\mathcal{S}=\mathcal{T}\times\mathcal{T}$). We can also naturally define the composition of special tree transformations: 
$$(T_1,T_2)\circ(T_2,T_3)=(T_1,T_3).$$

Given two special tree transformations $f,g\in\mathcal{S}$ where $f=(t_1,t_2), g=(T_1,T_2)$. Define $f$ is a \textbf{Local Depiction} of $g$, denoted as $f\preceq g$, if and only if: 

(1) $t_1$ and $t_2$ are respectively subtrees of $T_1$ and $T_2$: 
\[t_1\subseteq T_1\wedge t_2\subseteq T_2.\]

(2) $g$ can be depicted by $f$: 
\[T_1/t_1=T_2/t_2.\]

Define $f$ is a \textbf{Co-local Depiction} of $g$, denoted as $f\preccurlyeq g$, if and only if there exists a disjoint list of common subtrees $\tau_1,\tau_2,\dots,\tau_m$ of $T_1$ and $T_2$, such that: 
\[
\begin{cases}
    T_1/\tau_1,\dots,\tau_m = t_1\\
    T_2/\tau_1,\dots,\tau_m = t_2
\end{cases}
\]

We may naturally infer that the local depiction relation is a partial order. For every special tree transformation $f\in\mathcal{S}$, define the \textbf{Generalized Tree Transformation} of $f$, denoted as $\varphi_f$, as the collection of all special tree transformations that can be locally depicted by $f$: 
\[\varphi_f:=\{g|f\preceq g\}\cup\{g|f\preccurlyeq g\}\]

Denote the collection of all generalized tree transformations as $\mathcal{G}$: 
\[\mathcal{G}:=\{\varphi_f|f\in\mathcal{S}\}\]
\end{definition}

\smallsec{Formulation}
Let $\mathcal{H}$ be a disjoint subset of $\mathcal{G}$ that represents the set of allowed transformations, where every two general tree transformations do not intersect. Let $\mathcal{L}:\mathcal{H}\to\mathbb{N}$ be the cost function that assigns a specific cost value in $\mathbb{N}$ (or potentially other number systems depending on the demand) to every general tree transformation available, with $\text{Id}\mapsto 0$. The generalized tree edit distance $d_{\text{GTED}}$ between two operator trees $T_1$ and $T_2$, with available transformations in $\Phi$ and a cost function $\mathcal{L}$, is defined to be the infimum of total cost of all possible finite sequences of special tree transformations that convert $T_1$ to $T_2$: 
\[d_{\text{GTED}}(\mathcal{H},\mathcal{L};T_1, T_2):=\inf\left(\sum_{i=1}^n\mathcal{L}(\varphi_{f_i})\right),\]
where $f_1,\dots,f_n$ satisfy $T_2=f_n\circ\cdots\circ f_1(T_1)$ and $\varphi_{f_i}\in\mathcal{H}$. 


For the purpose of binary semantic evaluation, we map the continuous loss value to a discrete outcome. This is achieved by transforming the loss into a normalized similarity score on the interval [0,1] and subsequently applying a decision threshold $\theta$. The first formulation is analogous to the conventional TED similarity, and we refer the interested reader to the comprehensive survey by \citep{bille2005survey} for more details on the algorithm.
$$
\delta_{\text{GTED}}(\theta;T_1,T_2) := \mathbb{I}_{\{\cdot>\theta\}}\left( 1-\frac{d_{\text{GTED}}(\mathcal{H},\mathcal{L};T_1, T_2)}{\max(|T_1|,|T_2|)}\right), 
$$
where $|T|$ stands for the number of vertices in the tree $T$, and the overall expression serves as an indicator for non-identity, yielding 1 if the trees differ and 0 if they are the same.

In practice, $\mathcal{H}$ is likely to include: (1) ``legal'' transformations, which depict the effect of proof steps in Lean, with certain restrictions in order for better alignment with human intuition; (2) ``dumb'' transformations, like those in the original TED algorithm to guarantee a probably reasonable result when a formal proof is unattainable. When no ``dumb'' transformations are allowed, there may be no proper sequences available, in which case the total loss is $+\infty$, which in a binary evaluation process gives a negative result directly. The normalization process may also fail. 

Now the standard TED algorithm can be considered as a special case of our GTED metric if we let $\mathcal{G}=\{\text{Inserting}, \text{Deleting}, \text{Relabeling}\}$ and normally set $\mathcal{L}:=\varphi\mapsto 1$. So theoretically, $\mathcal{H}$, $\mathcal{L}$, and $\theta$ can be deliberately designed according to demand. 

\smallsec{$\alpha$-conversion}
To serve as an example for the generalized tree transformations, we choose one of the most reasonable operations in formal mathematics, renaming the names of variables or hypotheses, which corresponds to $\alpha$-conversion in the language of $\lambda$-calculus, one of the foundational theories of Lean. 
$\alpha$-conversion says that, renaming the same variable in a formula does not change its essence: 
\[\boxed{\lambda(x).f[x]\xlongequal{x\mapsto y}\lambda(y).f[y]}.\]

Besides regular $\lambda$-abstractions, numerous common operators like dependent function types ($\Pi_{(x:\alpha)}, \beta(x)$), universal and existential quantifiers ($\forall,\exists$) all behave similarly in Lean, since they're also defined with $\lambda$-abstractions. In actual syntax of $\lambda$-calculus, the name of a variable is overwritten if undergone another $\lambda$-abstraction, as demonstrated in the example below, where the $x$'s are considered different so that the inner $x$ overwrites the outer one, thus giving ${\color{blue}z}$ as the result instead of ${\color{red}y}$: 
\[\boxed{\lambda({\color{red}x}).(\lambda({\color{blue}x}).{\color{blue}x}){\color{red}y}{\color{blue}z}={\color{blue}z}}.\]

In Lean, an $\alpha$-conversion between two statements can be exemplified as follows (assuming \texttt{P} is a predefined predicate with type \texttt{Nat -> Prop}): 
\begin{lstlisting}[language=lean, numbers=left, numberstyle=\tiny]
theorem t1 (x : Nat) : P x := by sorry
theorem t2 (y : Nat) : P y := by sorry
\end{lstlisting}
Under the framework of GTED, $\alpha$-conversion can be rebuilt as a generalized tree transformation, which allows it to be handled as a formal proof step that rigorously preserves mathematical semantics.
\begin{figure}[h]
\centering
\[
    \xymatrix@C=1mm{
        & \boxed{\texttt{t1}} \ar[dl]\ar[dr]\ar@/^2em/@{-->}[rrrr]^{\text{$\alpha$-conversion}} & & & & \boxed{\texttt{t2}} \ar[dl]\ar[dr]\\
        \boxed{\texttt{\_:\_}} \ar[d]\ar[dr] & & \boxed{\texttt{P}}\ar[d] & &  \boxed{\texttt{\_:\_}} \ar[d]\ar[dr] & & \boxed{\texttt{P}}\ar[d]\\
        \boxed{\color{blue}\texttt{x}} & \boxed{\texttt{Nat}} & \boxed{\color{blue}\texttt{x}} & & \boxed{\color{red}\texttt{y}} & \boxed{\texttt{Nat}} & \boxed{\color{red}\texttt{y}}
    }
\]
\caption{An example of $\alpha$-conversion, the renaming of a bound variable.}
\label{fig:alpha_conversion}
\end{figure}
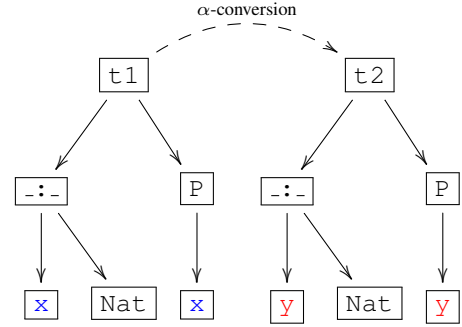

\begin{table*}[t]
\centering
\caption{Performance of automated evaluation metrics for statement formalization. Detailed results are available in Appendix \ref{app:detailed}.}
\label{tab:main_result}
\resizebox{1\textwidth}{!}{
\begin{tabular}{l|cccc|cccc}
\toprule[1pt]
\multirow{2}{*}{\textbf{Metric}} & \multicolumn{4}{c|}{\textbf{miniF2F}} & \multicolumn{4}{c}{\textbf{ProofNet}} \\ \cmidrule{2-9} 
& Precision & Recall & Accuracy & Kappa & Precision & Recall & Accuracy & Kappa\\ 
\midrule
Identity Match & \textbf{100.00\%} & 11.48\% & 47.32\% & 0.095 & 0/0 & 0.00\% & 47.31\% & 0.000 \\
Typecheck & 59.51\% & \textbf{100.00\%} & 59.51\% & 0.000 & 52.69\% & \textbf{100.00\%} & 52.69\% & 0.000 \\
BLEU & 78.22\% & \underline{64.75\%} & \underline{68.29\%} & 0.368 & 72.34\% & \underline{69.39\%} & \textbf{69.89\%} & 0.398 \\
Majority Voting & 88.00\% & 54.10\% & \underline{68.29\%} & 0.397 & \underline{78.38\%} & 59.18\% & \textbf{69.89\%} & \textbf{0.404} \\
Definitional Equality & \textbf{100.00\%} & 36.07\% & 61.95\% & 0.314 & 60.00\% & 6.12\% & 48.39\% & 0.015 \\
BEq & \underline{98.28\%} & 46.72\% & 67.80\% & \underline{0.405} & \textbf{100.00\%} & 16.33\% & \underline{55.91\%} & 0.156 \\
\cellcolor{cyan!20}\textbf{GTED} & \cellcolor{cyan!20}88.75\% & \cellcolor{cyan!20}58.20\% & \cellcolor{cyan!20}\textbf{70.73\%} & \cellcolor{cyan!20}\textbf{0.438} & \cellcolor{cyan!20}75.61\% & \cellcolor{cyan!20}63.27\% & \cellcolor{cyan!20}\textbf{69.89\%} & \cellcolor{cyan!20}\underline{0.402}\\
\bottomrule[1pt]
\end{tabular}
}
\end{table*}

\section{Experiments} \label{sec:experiments}
\subsection{Experiment Setting}
\smallsec{Dataset}
For evaluation, we use the miniF2F-test \citep{minif2f} and ProofNet-test \citep{proofnet} datasets. The specific versions employed in this study are sourced from Numina\footnote{https://huggingface.co/datasets/AI-MO/minif2f\_test} for miniF2F-test, and from DeepSeek\footnote{https://github.com/deepseek-ai/DeepSeek-Prover-V1.5/tree/main/datasets} for ProofNet-test. Moreover, since automated evaluation requires both ground truth and predicted formal statements, we employ HERALD Translator \citep{herald}, a state-of-the-art autoformalization model, to translate these datasets into their respective formal statements for subsequent evaluation.

\smallsec{Baseline Metrics}
We benchmark GTED against several competing baseline metrics to validate our proposed metric's efficacy. To ensure fair and consistent evaluation across these diverse metrics, we establish specific conventions for handling theorem names. For string-matching metrics (e.g., Identity Match, BLEU), we standardize theorem names in both ground truth and predicted formal statements to \texttt{thm}. Conversely, for proof-based metrics (e.g., BEq), we designate the ground truth theorem names as \texttt{thm\_P} and the predicted theorem names as \texttt{thm\_Q}. For all other metrics, theorem names remain unaltered.

\begin{itemize}[topsep=0pt, itemsep=0pt, parsep=0pt]
    \item Identity Match: This metric considers a predicted formal statement as correct if, after removing all whitespace, it is identical to the ground truth.
    \item Typecheck: A predicted formal statement is deemed correct if it successfully compiles.
    \item BLEU: We align with the ProofNet \citep{proofnet} for the computation.
    \item Majority Voting: Following the setup described in Lean Workbook \citep{lean_workbook} and BEq \citep{beq}, we employ DeepSeek-V3 \citep{deepseek_v3} with temperature 0.7 for 16 rounds of majority voting.
    \item Definitional Equality \citep{beq}: This metric directly attempts to prove \texttt{example: thm\_P = thm\_Q := by rfl}. If the proof is successful, the predicted formal statement is considered correct.
    \item BEq \citep{beq}: This metric involves attempting to prove \texttt{thm\_Q} using \texttt{thm\_P} and vice-versa, employing a broader range of tactics. A predicted formal statement is considered correct if both directions of proof are successfully completed. 
\end{itemize}

Regarding tactic usage for BEq, we adopt the ``Normal set'' of tactics \{\texttt{exact}, \texttt{exact?}, \texttt{have}, \texttt{apply}, \texttt{cases'}, \texttt{constructor}, \texttt{ext}, \texttt{intro}, \texttt{intros}, \texttt{rw}, \texttt{use}\}, as this set demonstrates the best performance in the original paper. Furthermore, we observe that a sampling configuration of temperature $T$=0.1 and top\_p=0.9 leads to more successful bidirectional proofs compared to the beam search used in the original paper (miniF2F: 58 vs 53, ProofNet: 8 vs 8). Consequently, we adopt this sampling setting for our experiments.

\smallsec{Human Evaluation}
To establish the ground truth labels, the miniF2F-test and ProofNet-test datasets are first translated using the HERALD Translator. From these translated results, 205 entries from miniF2F-test and 93 from ProofNet-test successfully compile. Four Lean4 human experts then judge the correctness of these predicted formal statements. Subsequently, different metrics are evaluated by comparing their results against these human-expert judgments, reporting Precision, Recall, Accuracy, and Kappa scores.

\smallsec{Implementation Details}
A fully-featured implementation of $\alpha$-conversion requires both variable renaming and scope-awareness to ensure consistency. Due to the time constraints of the present work, we have provisionally deferred the implementation of the scope-aware aspect and implemented only the primary renaming operation.

All experiments in this work are conducted using the Lean toolchain v4.19.0-rc2 on a single NVIDIA A100 GPU with 40GB of memory.

\begin{figure*}[t]
\centering
\includegraphics[width=2\columnwidth]{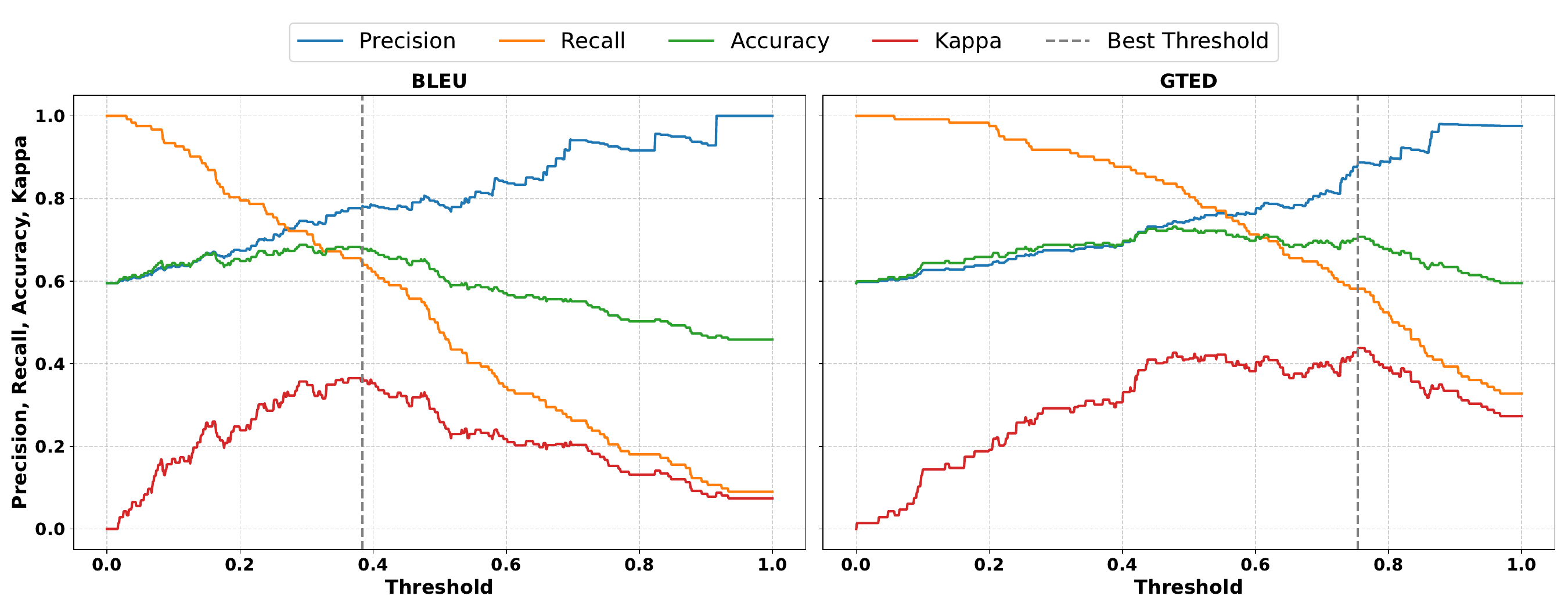}
\caption{Comparison of BLEU and GTED across thresholds on miniF2F.}
\label{fig:threshold_on_minif2f}
\end{figure*}

\subsection{Experiment Results}
\smallsec{Overall Comparison}
Table \ref{tab:main_result} presents a comprehensive overview of our proposed GTED metric compared to various baseline metrics across the miniF2F and ProofNet benchmarks. Our primary focus for evaluation is on \textbf{Kappa} and \textbf{accuracy}, as Kappa \citep{kappa} offers a robust measure of agreement beyond chance, and accuracy directly reflects the overall correctness of the evaluation. GTED demonstrates a superior ability to reliably assess formal statements, achieving the best overall performance. It achieves the highest Kappa score on miniF2F (\textbf{0.438}) and the second-highest on ProofNet (\textbf{0.402}), indicating a \textbf{moderate level of agreement} (0.4-0.6) with human expert evaluations across both datasets. This consistent performance underscores GTED's robustness, a quality not matched by other metrics, which often fail to maintain such agreement. Furthermore, GTED leads in accuracy (\textbf{70.73\%} on miniF2F and \textbf{69.89\%} on ProofNet).

\smallsec{Comparison with BLEU}
Both GTED and BLEU metrics produce continuous-valued outputs requiring thresholding for binary classification. While Table \ref{tab:main_result} reports their highest Kappa scores across all thresholds, Figure \ref{fig:threshold_on_minif2f} highlights a key advantage of GTED in terms of its \textbf{robustness to threshold selection}. BLEU's optimal Kappa is confined to a narrow threshold window, with slight deviations causing significant performance drops across all metrics. This sensitivity makes BLEU challenging to deploy reliably, as identifying and fixing the precise optimal threshold in practice is difficult. Conversely, GTED exhibits a robust performance plateau, maintaining high and consistent metric values over a much broader threshold range. This ``flatness'' renders GTED a more practical and dependable evaluation metric, a phenomenon similarly observed on ProofNet (see Appendix \ref{app:threshold}), reducing the burden of threshold selection and enhancing result generalizability.

\smallsec{Comparison with Majority Voting}
GTED provides a compelling alternative to LLM-based evaluation methods like Majority Voting. As shown in Table \ref{tab:main_result}, our approach is not only competitive but superior in performance, outperforming Majority Voting in both accuracy (e.g., 70.73\% vs. 68.29\%) and Kappa score (0.438 vs. 0.397) on miniF2F. Beyond its higher accuracy, our method operates at a significantly lower computational cost. This combination of being lightweight yet more reliable makes GTED highly advantageous in scenarios where inference efficiency and deployment simplicity are critical.

\smallsec{Comparison with Proof-based Metrics}
Comparing GTED with proof-based metrics like Definitional Equality and BEq, Table \ref{tab:main_result} reveals key distinctions. While Definitional Equality and BEq show strong Precision, their low Recall and Kappa scores, particularly on the more challenging ProofNet dataset, indicate an overly conservative approach that misses many correct formalizations. Moreover, the current development of automated theorem proving (ATP) exacerbates this issue. For instance, the SOTA proving model, DeepSeek-Prover-V2-671B \cite{deepseek_prover_v2}, currently achieves only a 7.45\% proof rate on the PutnamBench dataset \citep{putnambench}. This low proof rate implies that many essentially equivalent formalizations may not be provable by automated means, thereby leading to a high rate of false negatives. In contrast, GTED offers a more balanced, comprehensive, and robust assessment without being constrained by ATP's current development.

Additionally, we unexpectedly identify \textbf{three instances of false positives within the proof-based metrics}, which are further elaborated upon in Appendices \ref{app:fp_dee} and \ref{app:fp_beq}.

\section{Conclusion and Future Work}
We propose GTED (\textit{Generalized Tree Edit Distance}), a novel framework that assesses semantic similarity via the edit distance between operator trees of formal statements. Experiments on the miniF2F and ProofNet benchmarks show that GTED achieves state-of-the-art performance. It surpasses all baseline metrics on miniF2F and achieves top-tier results on ProofNet, including the joint-highest accuracy. These results demonstrate GTED's superior alignment with human expert judgment, providing the community with a more faithful metric for automated evaluation.

Our future work will focus on enhancing GTED by incorporating deeper mathematical domain knowledge. The immediate priority is to complete our implementation of $\alpha$-conversion by making it fully scope-aware, which will improve the handling of bound variables. More broadly, we aim to address the primary limitation of the current metric: its ``semantic naivety," which incorrectly penalizes logically equivalent yet syntactically different expressions (e.g., $x+y$ and $y+x$). To overcome this, we will develop a semantically-aware framework by augmenting GTED with a system of low-cost rewrite rules derived from fundamental mathematical axioms and theorems. These enhancements will ultimately transform GTED into a more robust and adaptable evaluation tool, capable of gauging true mathematical meaning beyond superficial syntactic structure.

\bibliography{main}
\bibliographystyle{icml2025}

\newpage
\appendix
\onecolumn
\section{Additional Experimental Analysis}
\subsection{Detailed Experimental Results} \label{app:detailed}
Tables \ref{tab:total_result_minif2f} and \ref{tab:total_result_proofnet} present the detailed experimental results, where TP, TN, FP, and FN denote the number of true positives, true negatives, false positives, and false negatives, respectively. Notably, BEq exhibits one false positive in the miniF2F dataset, while Definitional Equality also records two false positives in ProofNet.
\begin{table}[ht]
\centering
\caption{Detailed experimental results of automated evaluation metrics on miniF2F. }
\label{tab:total_result_minif2f}
\resizebox{0.9\textwidth}{!}{
\begin{tabular}{lcccccccc}
\toprule[1pt]
\multirow{2}{*}{\textbf{Metric}} & \multicolumn{8}{c}{\textbf{miniF2F}} \\ \cmidrule{2-9} 
& TP & TN & FP & FN & Precision & Recall & Accuracy & Kappa\\ 
\midrule
Identity Match & 14 & 83 & 0 & 108 & \textbf{100.00\%} & 11.48\% & 47.32\% & 0.095 \\
Typecheck & 122 & 0 & 83 & 0 & 59.51\% & \textbf{100.00\%} & 59.51\% & 0.000 \\
BLEU & 79 & 61 & 22 & 43 & 78.22\% & \underline{64.75\%} & \underline{68.29\%} & 0.368 \\
Majority Voting & 66 & 74 & 9 & 56 & 88.00\% & 54.10\% & \underline{68.29\%} & 0.397 \\
Definitional Equality & 44 & 83 & 0 & 78 & \textbf{100.00\%} & 36.07\% & 61.95\% & 0.314 \\
BEq & 57 & 82 & 1 & 65 & \underline{98.28\%} & 46.72\% & 67.80\% & \underline{0.405} \\
\cellcolor{cyan!20}\textbf{GTED} & \cellcolor{cyan!20}71 & \cellcolor{cyan!20}74 & \cellcolor{cyan!20}9 & \cellcolor{cyan!20}51 & \cellcolor{cyan!20}88.75\% & \cellcolor{cyan!20}58.20\% & \cellcolor{cyan!20}\textbf{70.73\%} & \cellcolor{cyan!20}\textbf{0.438}\\
\bottomrule[1pt]
\end{tabular}
}
\end{table}

\begin{table}[ht]
\centering
\caption{Detailed experimental results of automated evaluation metrics on ProofNet.}
\label{tab:total_result_proofnet}
\resizebox{0.9\textwidth}{!}{
\begin{tabular}{lcccccccc}
\toprule[1pt]
\multirow{2}{*}{\textbf{Metric}} & \multicolumn{8}{c}{\textbf{ProofNet}} \\ \cmidrule{2-9} 
& TP & TN & FP & FN & Precision & Recall & Accuracy & Kappa\\ 
\midrule
Identity Match & 0 & 44 & 0 & 49 & 0/0 & 0.00\% & 47.31\% & 0.000 \\
Typecheck & 49 & 0 & 44 & 0 & 52.69\% & \textbf{100.00\%} & 52.69\% & 0.000 \\
BLEU & 34 & 31 & 13 & 15 & 72.34\% & \underline{69.39\%} & \textbf{69.89\%} & 0.398 \\
Majority Voting & 29 & 36 & 8 & 20 & \underline{78.38\%} & 59.18\% & \textbf{69.89\%} & \textbf{0.404} \\
Definitional Equality & 3 & 42 & 2 & 46 & 60.00\% & 6.12\% & 48.39\% & 0.015 \\
BEq & 8 & 44 & 0 & 41 & \textbf{100.00\%} & 16.33\% & \underline{55.91\%} & 0.156 \\
\cellcolor{cyan!20}\textbf{GTED} & \cellcolor{cyan!20}31 & \cellcolor{cyan!20}34 & \cellcolor{cyan!20}10 & \cellcolor{cyan!20}18 & \cellcolor{cyan!20}75.61\% & \cellcolor{cyan!20}63.27\% & \cellcolor{cyan!20}\textbf{69.89\%} & \cellcolor{cyan!20}\underline{0.402} \\
\bottomrule[1pt]
\end{tabular}
}
\end{table}

\subsection{Two FP Cases for Definitional Equality} \label{app:fp_dee}
\begin{oframed}
\textbf{\# exercise\_1\_3\_8} \\
\textbf{NL: } Prove that if $\Omega=\{1,2,3, \ldots\}$ then $S_{\Omega}$ is an infinite group.\\

\textbf{Label: }
\vspace{-5pt}
\begin{lstlisting}[style=appendixstyle]
theorem thm_P : Infinite (Equiv.Perm ℕ) := by sorry
\end{lstlisting}
\vspace{-5pt}
\textbf{Prediction: }
\vspace{-5pt}
\begin{lstlisting}[style=appendixstyle]
theorem thm_Q : {Ω : Type u_1} [Infinite Ω] : Infinite (Equiv.Perm Ω) := by sorry
\end{lstlisting}
\vspace{-5pt}
\textbf{Definitional Equality}
\vspace{-5pt}
\begin{lstlisting}[style=appendixstyle]
example : thm_P = thm_Q := by rfl
\end{lstlisting}
\vspace{-5pt}
\end{oframed}
In Prediction, the original proposition is generalized, resulting in a semantic difference. However, in Definitional Equality, the Lean compiler will automatically infer that $\Omega$ in Prediction is the set of natural numbers $\mathbb{N}$, thus passing the verification and leading to a false positive.

\begin{oframed}
\textbf{\# exercise\_9\_4\_2c}\\
\textbf{NL: } Prove that $x^4+4x^3+6x^2+2x+1$ is irreducible in $\mathbb{Z}[x]$.\\

\textbf{Label: }
\vspace{-5pt}
\begin{lstlisting}[style=appendixstyle]
theorem thm_P : Irreducible  (X^4 + 4*X^3 + 6*X^2 + 2*X + 1 : Polynomial ℤ) := by sorry
\end{lstlisting}
\vspace{-5pt}
\textbf{Prediction: }
\vspace{-5pt}
\begin{lstlisting}[style=appendixstyle]
theorem thm_Q : Irreducible (wilsons_poly : ℤ[X]) := by sorry
\end{lstlisting}
\vspace{-5pt}
\textbf{Definitional Equality}
\vspace{-5pt}
\begin{lstlisting}[style=appendixstyle]
example : thm_P = thm_Q := by rfl
\end{lstlisting}
\vspace{-5pt}
\end{oframed}
In the Prediction, an undefined variable wilsons\_poly appeared instead of the polynomial given in the problem, resulting in a semantic difference. However, Lean will automatically interpret wilsons\_poly as an implicit variable. Therefore, in Definitional Equality, the Lean compiler will automatically infer wilsons\_poly as the polynomial $X^4 + 4X^3 + 6X^2 + 2X + 1$, thus passing the verification and leading to a false positive.

\subsection{One FP Case for BEq} \label{app:fp_beq}
\begin{oframed}
\textbf{\# mathd\_numbertheory\_254}\\
\textbf{NL: } Sally, Wei-Hwa, and Zoe are playing a game of marbles involving first arranging as many piles of 10 marbles as possible. Sally brought 239 marbles, Wei-Hwa brought 174 marbles, and Zoe brought 83 marbles. If all their marbles are grouped together, how many must be removed in order to start the game? Show that it is 6.\\

\textbf{Label: }
\vspace{-5pt}
\begin{lstlisting}[style=appendixstyle]
theorem thm_P : (239 + 174 + 83) % 10 = 6 := by sorry
\end{lstlisting}
\vspace{-5pt}
\textbf{Prediction: }
\vspace{-5pt}
\begin{lstlisting}[style=appendixstyle]
theorem thm_Q (s w z : ℕ) : (239 + 174 + 83) % 10 = 6 := by sorry
\end{lstlisting}
\vspace{-5pt}
\textbf{BEq: Label $\to$ Prediction}
\vspace{-5pt}
\begin{lstlisting}[style=appendixstyle]
theorem thm_Q (s w z : ℕ) : (239 + 174 + 83) % 10 = 6 := by
  exact thm_P
\end{lstlisting}
\vspace{-5pt}
\textbf{BEq: Prediction $\to$ Label}
\vspace{-5pt}
\begin{lstlisting}[style=appendixstyle]
theorem thm_P : (239 + 174 + 83) % 10 = 6 := by
  exact thm_Q 239 174 83
\end{lstlisting}
\vspace{-5pt}
\end{oframed}
The prediction is considered incorrect as it contains redundant variables \texttt{(s w z }$:\mathbb N$\texttt{)}. Human evaluators penalize such outputs, because the introduction of superfluous variables can cause confusion and represents an undesirable behavior for translator models.

\subsection{Performance of BLEU and GTED on ProofNet} \label{app:threshold}
\begin{figure*}[t]
\centering
\includegraphics[width=\columnwidth]{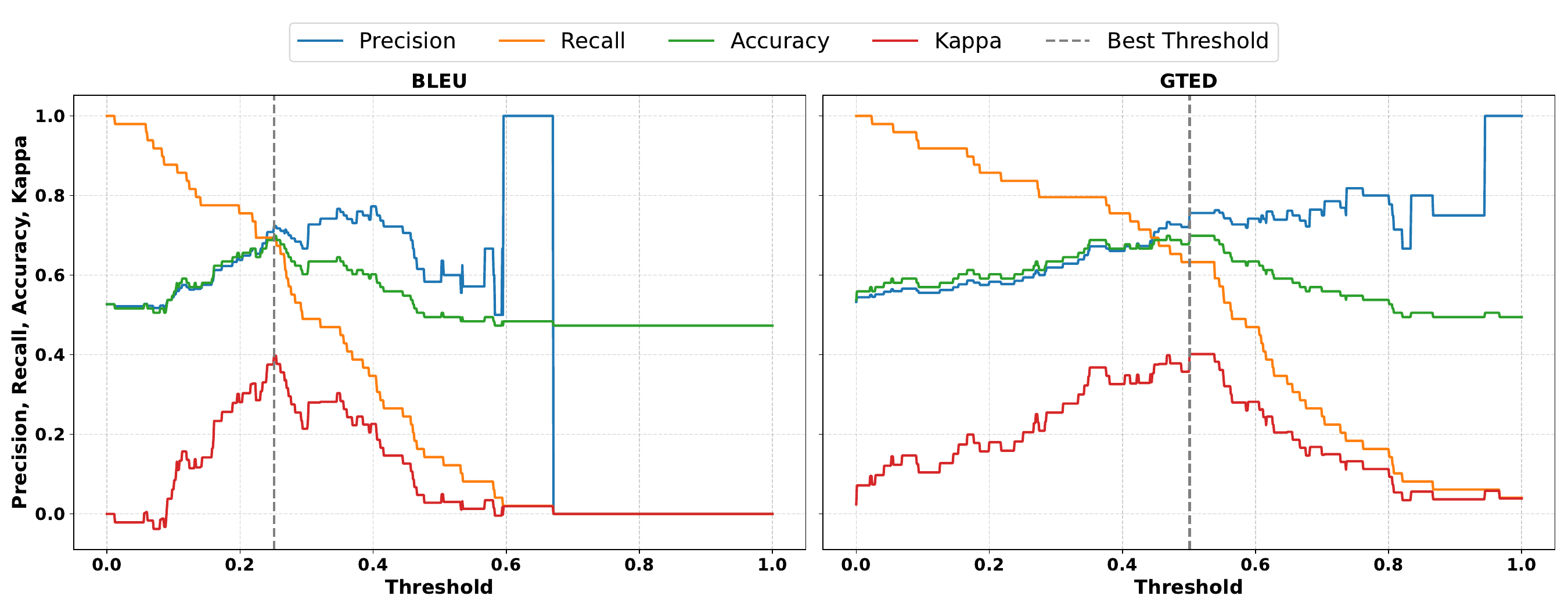}
\caption{Comparison of BLEU and GTED across thresholds on ProofNet.}
\label{fig:threshold_on_proofnet}
\end{figure*}

Figure \ref{fig:threshold_on_proofnet} shows the performance of BLEU and GTED across thresholds on ProofNet.

\section{Prompt Templates}
This section presents the prompts for the majority voting phase, which uses InternLM2-Math-Plus-7B \citep{internlm_math_plus} for back-translation and DeepSeek-V3 \citep{deepseek_v3} for NLI Check.

\begin{tcolorbox}[colframe=purple, width=1\linewidth, arc=1mm, auto outer arc, title={Prompt Template for Back-Translation}]
[UNUSED\_TOKEN\_146]user\textbackslash nConvert the formal statement into natural language:\textbackslash n```
lean\textbackslash n{formal\_statement}\textbackslash n```[UNUSED\_TOKEN\_145]\textbackslash n[UNUSED\_TOKEN\_146]assistant\textbackslash n
\end{tcolorbox}

\begin{tcolorbox}[colframe=purple, width=1\linewidth, arc=1mm, auto outer arc, title={Prompt Template for NLI Check}]
Please check following two math problems is same or different? Please consider each statement in two problems, they are different if any statement is different. Please point out any differences you found. Please reply **same** or **different** in the final sentence with bold format.

Problem 1: \{THM\_1\}

Problem 2: \{THM\_2\}
\end{tcolorbox}

\end{document}